\documentclass{article} 
\usepackage{nips15submit_e,times}
\usepackage{hyperref}
\usepackage{url}
\usepackage{authblk}
\usepackage{amsmath,graphicx,booktabs}
\usepackage{epstopdf,amssymb}
\usepackage{multirow}
\usepackage{fixltx2e}
\usepackage{subfigure}
\usepackage{float}
\usepackage{titlesec}
\usepackage{bm}

\title{\textbf{MEDICAL IMAGE SUPER-RESOLUTION USING A GENERATIVE ADVERSARIAL NETWORK}} 
\author{Yongpei Zhu$^{1}$, Xuesheng Zhang$^{1}$, Kehong Yuan$^{1*}$\\
	$^{1}$Graduate School at Shenzhen, Tsinghua University, Shenzhen 518055, China.\\
	*Corresponding author: Kehong Yuan (e-mail: yuankh@sz.tsinghua.edu.cn)\\
	\texttt{zhuyp17@mails.tsinghua.edu.cn}\\
}

\nipsfinalcopy 

\begin{document}
	
	\maketitle
	
	\begin{abstract}
		During the growing popularity of electronic medical records, electronic medical record (EMR) data has exploded increasingly. It is very meaningful to retrieve high quality EMR in mass data. In this paper, an EMR value network with retrieval function is constructed by taking stroke disease as the research object. It mainly includes: 1) It establishes the electronic medical record database and corresponding stroke knowledge graph. 2) The strategy of similarity measurement is included three parts(patients' chief complaint, pathology results and medical images). Patients' chief complaints are text data, mainly describing patients' symptoms and expressed in words or phrases, and patients' chief complaints are input in independent tick of various symptoms. The data of the pathology results is a structured and digitized expression, so the input method is the same as the patient's chief complaint; Image similarity adopts content-based image retrieval(CBIR) technology. 3) The analytic hierarchy process (AHP) is used to establish the weights of the three types of data and then synthesize them into an indicator. The accuracy rate of similarity in top 5 was more than 85\% based on EMR database with more 200 stroke records using leave-one-out method. It will be the good tool for assistant diagnosis and doctor training, as good quality records are colleted into the databases, like Doctor Watson, in the future.
		
	\end{abstract}

\section{Method}

\subsection{Deformation Method}
Here, we introduce the deformation method that can generate JD and CV by the grid generation, which will be used in our later methods. Diffeomorphism is an active research topic in differential geometry [13]. JD and CV play an important role in determining a diffeomorphism. Consider $\mathrm{\Omega}$ and $\mathrm{\Omega}_t \subset \mathbb R^{2,3}$ with $0\leq{t}\leq{1}$, be moving (includes fixed) domains. Let $\pmb{v}(\pmb{x},t)$ be the velocity field on $\partial\mathrm{\Omega_t}$, where $\pmb{v}(\pmb{x},t)\cdot{\pmb{\mathrm{n}}}=0$ on any part of $\partial\mathrm{\Omega}_t$ with slippery-wall boundary conditions where $\pmb{\mathrm{n}}$ is the outward normal vector of $\partial\mathrm{\Omega}_t$. Given diffeomorphism $\pmb{\varphi}_0:\mathrm{\Omega}\rightarrow\mathrm{\Omega}_0$ and scalar function $f(\pmb{x},t)>0 \in C^1(\pmb{x},t)$ on the domain $\mathrm{\Omega}_t \times [0,1]$, such that

\begin{equation}\label{General1}
\begin{aligned}
&f(\pmb{x},0)=J(\pmb{\varphi}_0)\\
&\int_{\mathrm{\Omega}_t} \dfrac{1}{f(\pmb{x},t)}d\pmb{x} = |\mathrm{\Omega}_0|.
\end{aligned}
\end{equation}
A new (differ from $\pmb{\varphi}_0$) diffeomorphism $\pmb{\phi}(\pmb{\xi},t):\mathrm{\Omega}_0\rightarrow\mathrm{\Omega}_t$, such that  $J(\pmb{\phi}(\pmb{\xi},t)) =\text{det}\nabla(\pmb{\phi}(\pmb{\xi},t)) = f(\pmb{\phi}(\pmb{\xi},t),t)$, $\forall t \in [0,1]$,
can be constructed the following two steps:
\begin{itemize}
	\item First, determine $\pmb{u}(\pmb{x},t)$ on $\mathrm{\Omega}_t$ by solving[19]
	\begin{equation}\label{General2}
	\left\{
	\begin{aligned}
	\text{div } \pmb{u}(\pmb{x},t)& = -\frac{\partial}{\partial t}(\dfrac{1}{f(\pmb{x},t)}) \\
	\text{curl } \pmb{u}(\pmb{x},t)& = 0\\
	\pmb{u}(\pmb{x},t)& = \dfrac{\pmb{v}(\pmb{x},t)}{f(\pmb{x},t)} \text{, on } \partial\mathrm{\Omega}_t
	\end{aligned}\right.
	\end{equation}
	\item Second, determine $\pmb{\phi}(\pmb{\xi},t)$ on $\mathrm{\Omega}_0$ by solving	
	\begin{equation}\label{General3}
	\left\{
	\begin{aligned}
	\frac{\partial \pmb{\phi}(\pmb{\xi},t)}{\partial t}& = f(\pmb{\phi}(\pmb{\xi},t),t) \pmb{u}(\pmb{\phi}(\pmb{\xi},t),t), \\
	\pmb{\phi}(\pmb{\xi},0)& = \pmb{\varphi}_{0}(\pmb{\xi})
	\end{aligned}\right.
	\end{equation}
\end{itemize}

For computational simplicity system (\ref{General2}) is modified into a Poisson equation as follows. Let $\pmb{u}(\pmb{x},t)= \nabla \pmb{w}(\pmb{x},t)$, then
\begin{equation}\label{General4}
\Delta \pmb{w}(\pmb{x},t) = \text{div } \nabla \pmb{w}(\pmb{x},t) = \text{div } \pmb{u}(\pmb{x},t) = -\frac{\partial}{\partial t}(\dfrac{1}{f(\pmb{x},t)})
\end{equation}

\section{Loss Function}
Here, we formulate the perceptual loss as the weighted sum of a content loss(${l}_{X}^{SR}$) and an adversarial loss component as:
\begin{equation}
{l}^{SR}={l}_{X}^{SR}+{10}^{-3}{l}_{Gen}^{SR}
\end{equation}

We replace the loss calculated on feature maps of VGG[17]
with a loss calculated on CV feature maps of reconstructed image G(${I}^{LR}$) and the reference image ${I}^{HR}$, which are more invariant to changes in pixel space. We define the content loss ${l}_{X}^{SR}$ as the Euclidean distance between the CV feature information of a reconstructed image G(${I}^{LR}$) and the reference image ${I}^{HR}$:
\begin{equation}
{l}_{CV}^{SR}=\frac{1}{{W}_{i,j}{H}_{i,j}}\sum\limits_{x=1}^{W_{i,j}}\sum\limits_{y=1}^{H_{i,j}}(CV({I}^{HR})_{x,y}-CV(G(y,{I}^{LR}))_{x,y})^{2}
\end{equation}
Here ${W}_{i,j}$ and ${H}_{i,j}$ describe the dimensions of the respective CV feature maps of ${I}^{HR}$ and $G(y,{I}^{LR})$. According to manifold learning, the geometric invariance of manifold plays an important role in improving image resolution, and CV feature map can better maintain the geometric invariance of manifolds, thus contributing to the optimization of image resolution.
And the adversarial(generative) loss ${l}_{Gen}^{SR}$ is defined based on the probabilities of the discriminator $D(y,G(y,{I}^{LR}))$ over all training samples as:
\begin{equation}
{l}_{Gen}^{SR}=\sum\limits_{n=1}^{N}{-logD(y,G(y,{I}^{LR}))})
\end{equation}
Here, $D(y,G(y,{I}^{LR}))$ is the probability that the reconstructed image  $G(y,{I}^{LR})$ is a natural HR image.
\begin{figure}[ht]
\centering
\includegraphics[scale=1,width=0.5\textwidth]{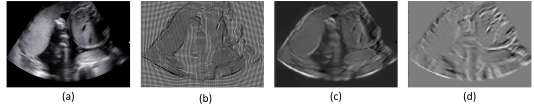}
\caption{Generated images based on JD and CV: (a) The original ultrasound image, (b) The grid image, (c) The image formed by JD, (d) The image formed by CV.}
\end{figure}

\section{EXPERIMENTS}

\subsection{Datasets and Evaluation Criteria}
\subsubsection{Datasets}
Since we do not have enough high-resolution ultrasound datasets, we used the face image dataset to train the model and then tested on the low-resolution ultrasound images. We validate our method on two datasets including CelebA[3] and ultrasound image.

\textbf{CelebFaces Attributes Dataset (CelebA)} is a large-scale face attributes dataset with more than 200K celebrity images, each with 40 attribute annotations. The images in this dataset cover large pose variations and background clutter. CelebA has large diversities, large quantities, and rich annotations, including 10,177 number of identities, 202,599 number of face images, and 5 landmark locations, 40 binary attributes annotatiper image.

\textbf{Ultrasound image dataset}: we used 1000 low-resolution ultrasound images from the clinic to test and evaluate the model.
\subsubsection{Evaluation Criteria}
\textbf{PSNR} Peak signal-to-noise ratio (PSNR) is a common objective measure used to measure the reconstruction quality of lossy transformations. PSNR is inversely proportional to the logarithm of the mean square error (MSE) between the real image and the generated image. It can be defined as:
\begin{equation}
\begin{aligned}
&MSE({I}^{HR},{I}^{SR})=\frac{1}{N}\sum\limits_{N}({I}^{HR}-{I}^{SR})^{2}\\
&PSNR=10lg(\frac{L^2}{MSE})
\end{aligned}
\end{equation}

In the above formula, L is the maximum possible pixel value (for 8-bit RGB images, it is 255).
\textbf{SSIM} Structural similarity (SSIM) is a subjective measure
used to measure the structural similarity between images based on three relatively independent comparisons (i.e., brightness, contrast, and structure).
\begin{equation}
SSIM({I}^{HR},{I}^{SR})=[C_{l}({I}^{HR},{I}^{SR})]^{\alpha}[C_{c}({I}^{HR},{I}^{SR})]^{\beta}[C_{s}({I}^{HR},{I}^{SR})]^{\gamma}
\end{equation}

In the formula above, ${\alpha,\beta}$ and ${\gamma}$ are weights of brightness, contrast, and structural-comparison functions, respectively.
The common expression of SSIM formula is as follows:
\begin{equation}
SSIM({I}^{HR},{I}^{SR})=\frac{(2{\mu}_{I^{HR}}{\mu}_{I^{SR}}+c_{1})({\sigma}_{I^{HR}I^{SR}}+c_{2})}{({\mu}_{I^{HR}}^{2}+{\mu}_{I^{SR}}^{2}+c_{1})({\sigma}_{I^{HR}}^{2}+{\sigma}_{I^{SR}}^{2}+c_{2})}
\end{equation}

${\mu}_{I}$ represents the average value of a particular image, and ${\sigma}_{I}$ represents the standard deviation of a particular image. ${\sigma}_{I\hat{I}}$ represents the covariance of two images. Since the statistical features or distortion of the image may be unevenly distributed, it is more reliable to evaluate the image quality locally than to apply the image quality globally. Mean SSIM is a local quality evaluation method, which divides the image into multiple windows and averages the SSIM obtained by each window.
\textbf{MOS} Mean Opinion Score is the most representative subjective evaluation method of quality. It judges the image quality through the normalization of the observer's rating. The higher the value, the better the subjective quality of the image.

\subsubsection{Implementation}
Our implementation uses Keras[18] with a Tensorflow backend[20] and the Adam optimizer[21] with a learning rate of $2\times{10}^{-4}$. We used MATLAB to generate the images formed by JD and CV information based on CelebA and ultrasound image, and saved it as jpg format. We set the epochs as 3000, batch size as 10, steps of per epoch as 100 using one GeForce GTX 1080 Ti GPU.

\begin{figure}[ht]
\centering
\includegraphics[scale=1,width=0.5\textwidth]{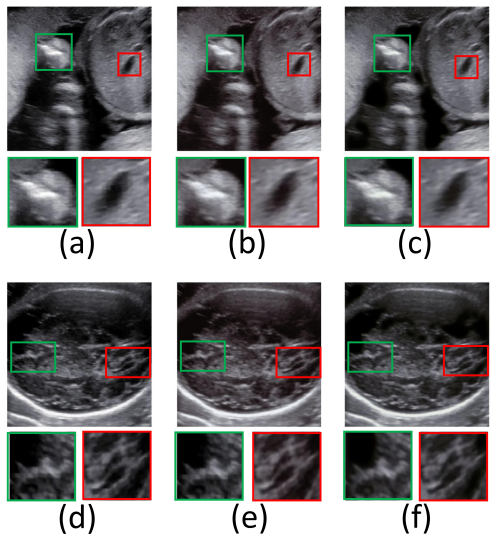}
\caption{Corresponding reference HR image and reconstruction results of SRGAN and our methods: (a) (d) Original HR image, (b)(e) SRGAN, (c) (f) ours.}
\end{figure}

\section{Results}
\subsubsection{Performance of CSRGAN}
In this section, we evaluate our method CSRGAN with the same VGG loss $l_{VGG/5,4}^{SR}$. Quantitative results are summarized in Table 1 and visual examples provided in Fig. 3. We conducted a MOS test to quantify the ability of different methods to reconstruct perceptually convincing images. Specifically, we asked 5 raters to assign an integral score from 1 (bad quality) to 5 (excellent quality) to the super- resolved images on 3 versions of each image on ultrasound image dataset: the original HR image, SRGAN and our method CSRGAN. In Fig.3 we can see that our method yields better texture detail when compared to SRGAN. This confirm that our method CSRGAN(in terms of PSNR/SSIM/MOS in Table 1) significantly outperformed SRGAN on the dataset and sets a new state of the art on the dataset.
\subsubsection{Performance of proposed content loss}
We investigated the effect of different content loss choices in the perceptual loss for the GAN-based networks. Quantitative
results are summarized in Table 2. We can see that CSRGAN-CV significantly outperformed other CSRGAN and SRGAN variants on the dataset. This represents our method can make full use of the idea of using GANs to learn manifold features better such as JD and CV, which plays an important role in manifold learning.

\begin{table}[H]
	\begin{center}
		\begin{tabular}{|*{3}{c|}}
			\toprule
			\multicolumn{1}{|c|}{Method} & \multicolumn{1}{|c|}{SRGAN} & \multicolumn{1}{|c|}{ours}
			\\
			
			\midrule
			PSNR & 35.82 & $\mathbf{36.79}$\\
			
			
			SSIM & 0.9673 & $\mathbf{0.9701}$\\
			
			MOS & 3.57 & $\mathbf{3.78}$\\

			\bottomrule
		\end{tabular}
	\end{center}
	
	\caption{Summary and comparison of SRGAN-54 and CSRGAN-54 on test set: mean PSNR(dB), mean SSIM and mean MOS (higher is better).}
\end{table}

\begin{table}[H]
	\begin{center}
		\begin{tabular}{|*{8}{c|}}
			\toprule
			\multicolumn{1}{|c|}{Experiment} & \multicolumn{3}{|c|}{SRGAN-} & \multicolumn{4}{|c|}{CSRGAN-}\\
			\cmidrule{2-8}
			& MSE & VGG22 & VGG54 & MSE & VGG22 & VGG54 & CV \\
			\midrule
			PSNR & 36.54 & 35.87 & 35.82 & 37.43 & 36.84 & 36.79 & $\mathbf{37.68}$\\
			
			SSIM & 0.9706 & 0.9654 & 0.9673 & 0.9774 & 0.9689 & 0.9701 & $\mathbf{0.9784}$\\
			
			MOS & 3.52 & 3.54 & 3.57 & 3.67 & 3.76 & 3.78 & $\mathbf{3.83}$ \\

			\bottomrule
		\end{tabular}
	\end{center}
	
	\caption{Performance of different loss functions for SRGAN and CSRAGN on test set: mean PSNR, mean SSIM and mean MOS (higher is better).}
\end{table}

	\textbf{Acknowledgments}
	
	The authors would like to thank the project "Three Medical and Health Engineering of Shenzhen"(No. SZSM201811094).\\


\begin{thebibliography}{1}
	\providecommand{\url}[1]{#1}
	\csname url@samestyle\endcsname
	\providecommand{\newblock}{\relax}
	\providecommand{\bibinfo}[2]{#2}
	\providecommand{\BIBentrySTDinterwordspacing}{\spaceskip=0pt\relax}
	\providecommand{\BIBentryALTinterwordstretchfactor}{4}
	\providecommand{\BIBentryALTinterwordspacing}{\spaceskip=\fontdimen2\font plus
		\BIBentryALTinterwordstretchfactor\fontdimen3\font minus
		\fontdimen4\font\relax}
	\providecommand{\BIBforeignlanguage}[2]{{%
			\expandafter\ifx\csname l@#1\endcsname\relax
			\typeout{** WARNING: IEEEtranS.bst: No hyphenation pattern has been}%
			\typeout{** loaded for the language `#1'. Using the pattern for}%
			\typeout{** the default language instead.}%
			\else
			\language=\csname l@#1\endcsname
			\fi
			#2}}
	\providecommand{\BIBdecl}{\relax}
	\BIBdecl
	
	\bibitem{[1]}
	~C.Y. Yang, ~C. Ma, \& ~M.H. Yang. Single-image super-resolution: A benchmark. \emph{In European Conference on Computer Vision (ECCV)}. pp.372--386, 2014.1.
	
	\bibitem{[2]}
	~Y.P. Zhu, ~Z.C. Zhou, ~G.J. Liao, ~Q.X. Yang, ~K.H. Yuan.
	Effects of Differential Geometry Parameters on Grid Generation and Segmentation of MRI Brain Image. \emph{IEEE Access}.\textbf{7}(1),68529--68539(2019)
	
	\bibitem{[3]}
	~Z.W. Liu, ~P. Luo, ~X.G. Wang and ~X.O. Tang. Deep Learning Face Attributes in the Wild.  In IEEE International Conference on Computer Vision (ICCV),2015.12.
	
	\bibitem{[4]}
	~W. T. Freeman, ~E. C. Pasztor, and ~O. T. Carmichael. Learning low-level vision. International Journal of Computer Vision, vol. 40, no. 1, pp. 25--47, 2000. 2

	
\end{thebibliography}
\end{document}